\documentclass[journal]{IEEEtran}
\usepackage{amsmath,amsfonts}
\usepackage{array}
\usepackage[caption=false,font=normalsize,labelfont=sf,textfont=sf]{subfig}
\usepackage{textcomp}
\usepackage{stfloats}
\usepackage{url}
\usepackage{verbatim}
\usepackage{graphicx}
\usepackage{cite}

\usepackage{algorithm2e}
\usepackage{adjustbox}
\usepackage{multicol}
\usepackage{multirow}
\usepackage[dvipsnames]{xcolor}
\begin{document}

\title{Prediction Calibration for Generalized Few-shot Semantic Segmentation}

\author{Zhihe Lu, Sen He, Da Li, Yi-Zhe Song, \IEEEmembership{Senior Member,~IEEE}, Tao Xiang
\thanks{Manuscript received April 19, 2021; revised August 16, 2021.}
\thanks{Zhihe Lu (e-mail: zhihelu.academic@gmail.com), Yi-Zhe Song and Tao Xiang are with the Centre for Vision Speech and Signal Processing (CVSSP), University of Surrey, and also with iFlyTek-Surrey Joint Research Center on Artificial Intelligence, University of Surrey, Guildford GU2 7XH, United Kingdom.}
\thanks{Sen He is with Meta AI, London, United Kingdom.}
\thanks{Da Li is with Samsung AI center, Cambridge CB1 2JH, United Kingdom.}
}

\markboth{Journal of \LaTeX\ Class Files,~Vol.~14, No.~8, August~2021}%
{Shell \MakeLowercase{\textit{et al.}}: A Sample Article Using IEEEtran.cls for IEEE Journals}


\maketitle

\begin{abstract}
 Generalized Few-shot Semantic Segmentation (\textit{GFSS}) aims to segment each image pixel into either base classes with abundant training examples or novel classes with only a handful of (e.g., 1-5) training images per class. Compared to the widely studied Few-shot Semantic Segmentation (\textit{FSS}), which is limited to segmenting novel classes only, \textit{GFSS} is much under-studied despite being more practical. Existing approach to \textit{GFSS} is based on classifier parameter fusion whereby a  newly trained novel class classifier and a pre-trained base class classifier are combined to form a new classifier. As the training data is dominated by base classes, this approach is inevitably biased towards the base classes. In this work, we propose a novel Prediction Calibration Network (\textit{PCN}) to address this problem. Instead of fusing the classifier parameters, we fuse the scores produced separately by the base and novel classifiers. To ensure that the fused scores are not biased to either the base or novel classes, a new Transformer-based calibration module is introduced. It is known that the lower-level features are useful of detecting edge information in an input image than higher-level features. Thus, we build a cross-attention module that guides the classifier's final prediction using the fused multi-level features. However, transformers are computationally demanding. 
 Crucially, to make the proposed cross-attention module training tractable at the pixel level, this module is designed based on feature-score cross-covariance and episodically trained to be generalizable at inference time. Extensive experiments on PASCAL-$5^{i}$ and COCO-$20^{i}$ show that our \textit{PCN} outperforms the state-the-the-art alternatives by large margins.
\end{abstract}


\section{Introduction}
\label{sec:intro}

Image semantic segmentation has been studied for decades. In the past 8-10 years, it has made arguably the largest advancements, thanks to the availability of large-scale annotated datasets and advanced deep segmentation models \cite{chen2017deeplab,zhao2017pyramid,chen2018encoder}.
However, existing semantic segmentation methods are still limited in  real-world applications because of the small number of object classes they can recognize. This is due to the fact that data annotation for semantic segmentation is at the pixel level and thus particularly laborious and expensive.
For example, it took more than 70,000 worker hours to build the popular COCO \cite{lin2014microsoft} segmentation dataset with merely 80 common object classes. It is thus infeasible to scale the efforts to the millions of object classes existing in the natural world. 

Few-shot Semantic Segmentation (\textit{FSS}) offers a solution to the scalabilty problem and has been studied extensively in recent years \cite{shaban2017one,wang2019panet,nguyen2019feature,wangfew,yang2020prototype,zhang2019canet,zhang2020sg,lu2021simpler}. With \textit{FSS}, a new object class can be segmented with only a handful of training examples. To achieve this, existing methods follow the meta-learning paradigm that learns to learn a model by simulating the few-shot scenarios on a large-scale annotated dataset. 
The learned model can then adapt quickly to novel classes using only a few support examples per class. However, the existing \textit{FSS} setting is still unsuited for real-world deployments. This is because it focuses only on segmenting novel classes while ignoring the base classes that typically co-exist with the novel ones in the same scene. 

In contrast, a Generalized Few-shot Semantic Segmentation (\textit{GFSS}) setting is more practical and thus the focus of this work. 
A \textit{GFSS} model, once trained on abundant base class examples and a few novel class ones, is able to recognize both types of classes during inference time. Though it is clearly more practical, \textit{GFSS} has received very little attention with  \cite{liu2020dynamic,tian2020generalized} being only two existing works as far as we know. This is because \textit{GFSS} is also much more challenging than \textit{FSS} due to a base class bias problem. More specifically, a straightforward solution for \textit{GFSS}, adopted in \cite{liu2020dynamic,tian2020generalized}, is classifier parameter fusion. That is, one first learns a base classifier on the base training set, and then learns a novel classifier on the support set of novel classes via existing \textit{FSS} methods such as prototypical learning~\cite{dong2018few,liu2020part,wang2019panet}. The two classifiers are then fused into one to segment both base and novel classes. 
However, since the base classes have much more classes as well as annotated training examples per class, the fused classifier would be inevitably biased towards them (see results in Table \ref{tab:score_fusion} and Figure \ref{fig:joint_cls_pred}).

\begin{figure*}[ht]
    \centering
    \includegraphics[width=0.9\textwidth]{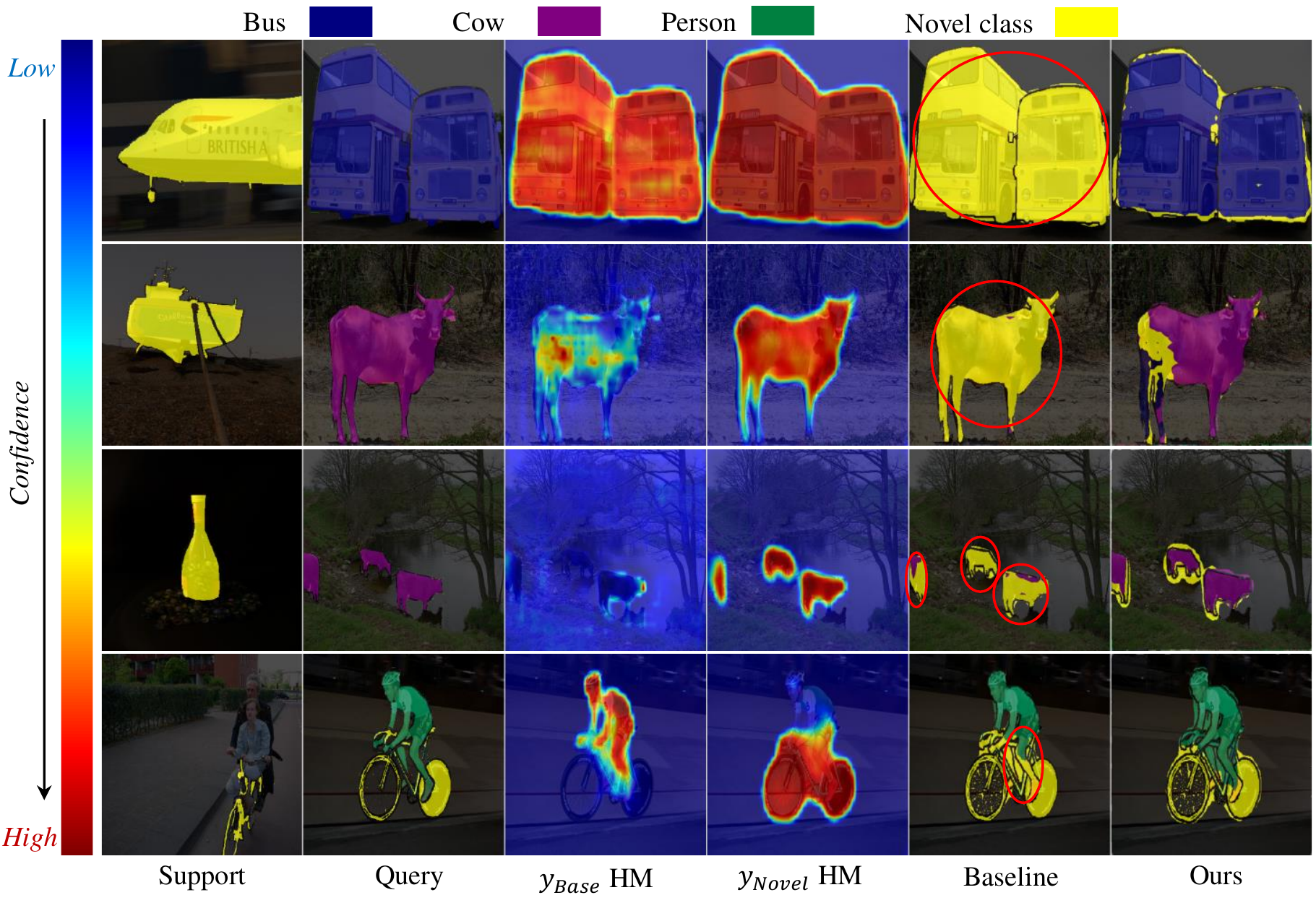}
    \caption{Illustrating the overconfidence problem under our Normalized Score Fusion (NSF). From left to right: support image, query image, base class classifier's prediction score heat map ($y_{base}$ HM) for chosen class (e.g., bus, cow and person), novel class classifier's prediction score heat map ($y_{novel}$ HM), NSF's result and our final model \textit{PCN}'s result. Each row contains one task (airplane, boat, bottle and bike as novel respectively). The red circles on NSF's result highlight the incorrect predictions caused by the overconfidence of NSF on novel classes -- base class pixels are  wrongly classified as novel.}
    \label{fig:visacm}
\end{figure*}

In this paper, we propose a novel Prediction Calibration Network (\textit{PCN}) for \textit{GFSS} to address the base class bias problem. Instead of classifier parameter fusion, we resort to classifier score fusion  with base-novel score calibration. Specifically, \textit{PCN} consists of two different calibration modules: Normalized Score Fusion (NSF) and Transformer based prediction calibration. NSF is designed to prevent base class bias. Concretely, two sets of prediction scores produced separately by the base and novel classifiers are normalized (by softmax) and then fused. We show empirically that by simply switching to score-level fusion, we can indeed effectively remove the base class bias (see Table \ref{tab:score_fusion}) and achieve the SOTA performance (see Table \ref{tab:coco_main}). 
However, another type of bias now appears, namely novel class bias (see Figure~\ref{fig:visacm}). This is because typically the number of novel classes is much smaller than base. A softmax based normalization for them is thus prone to overshooting their prediction confidence. A more principled approach to avoid both types of bias is thus needed. 

To that end, we propose to exploit the  cross attention between the model predictions (scores) and input features to enforce feature-score consistency, i.e., two pixels with similar features should be assigned to the same class. 
Especially, the features are designed to contain both detailed and semantic knowledge, thereby benefiting the calibration process by enriching the few-shot learned poor representation.
The calibration is implemented by a Transformer but it must be specially designed.
This is because a standard token (each pixel/patch regarded as a token) based cross attention is not applicable to our case. First, the pixel number is too large leading to an intractable computational complexity. Also the varying class numbers between the base training and final test require different projection heads. Therefore, inspired by the latest efficient Transformer design, cross-covariance Transformer~\cite{el2021xcit}, we propose a novel covariance based Transformer to tackle our problem.
More specifically, we treat each class (score) and feature channel as a token. After being embedded by the query and key projection heads separately, their cross-covariance is calculated to transform the value embeddings, which are mapped from the feature channel tokens.
Subsequently, the transformed values are mapped by a calibration head to calibrate the original prediction scores. 
Our Transformer is episodically trained so that it can generalize during inference time. 

Our {\bf contributions} are as follows. 
\begin{itemize}
    \item For the first time, we found and investigated the core issue of existing \textit{GFSS} methods. i.e., base class bias issue, due to directly combining two types of classifiers' parameters, and proposed a simple yet efficient baseline -- Normalized Score Fusion (NSF). NSF alone can prevent the base-class bias problem caused by classifier-parameter-level fusion, but is prone to novel-class bias.
    \item To address the novel-class bias, a novel feature-score cross-covariance based Transformer is further proposed and meta-learned for prediction calibration. Crucially, we introduce the multi-level fused feature for the few-shot learned poor representation enrichment.
    \item Extensive experiments on PASCAL-$5^{i}$ and COCO-$20^{i}$ demonstrate that our \textit{PCN} (NSF and Transformer combined) outperforms the state-of-the-art alternatives by significant margins.
\end{itemize}

\begin{figure*}[ht]
    \centering
    \includegraphics[width=1.0\textwidth]{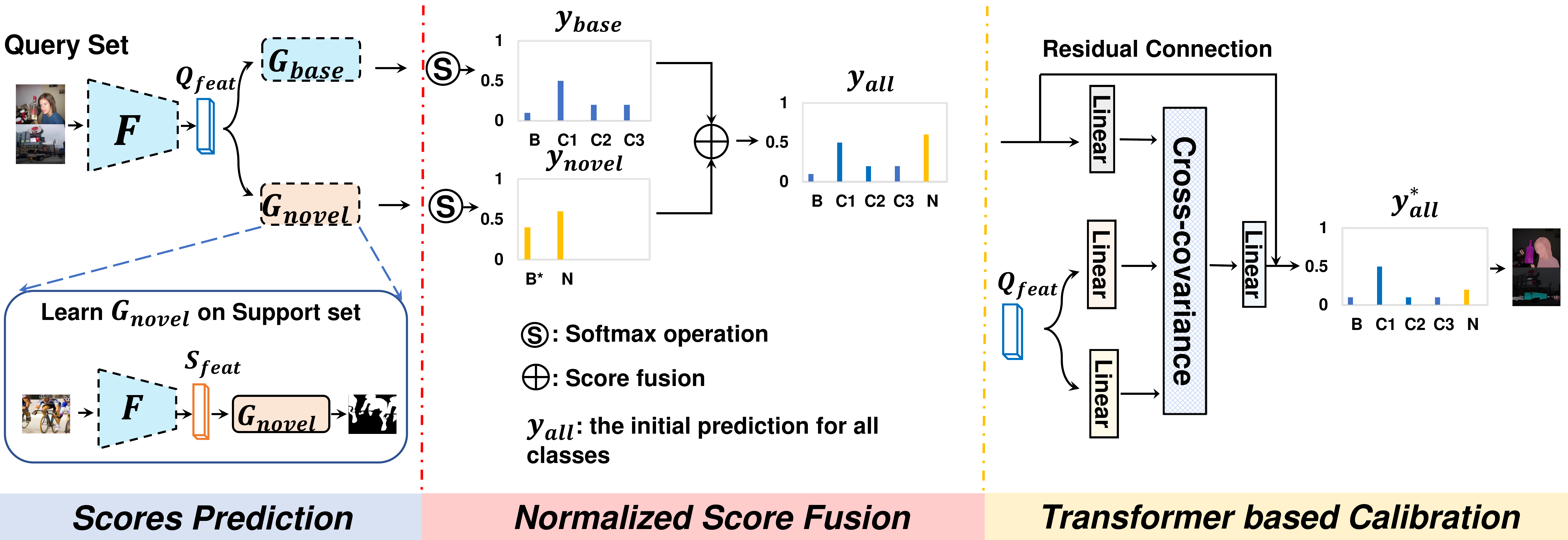}
    \caption{An overview of our proposed pipeline. 
    A feature extractor $F$ and a base classifier $G_{base}$ are first trained by standard segmentation training on base classes (not shown in this figure) and frozen in subsequent training stages.
    In the meta-training stage, we first learn a novel classifier $G_{novel}$ on a support set and fix its parameters.
    With $G_{base}$ and $G_{novel}$, we can now obtain two prediction scores $y_{base}$ and $y_{novel}$ on a query set.
    Normalized Score Fusion (NSF) is then applied on $y_{base}$ and $y_{novel}$ to form the initial prediction scores $y_{all}$ for all classes.
    Finally, $y_{all}$ is calibrated by a Transformer-based calibration module for the final prediction. 
    During inference, given a new task, only a novel classifier is needed to be trained on a support set while other modules are frozen.
    Note that the prediction scores shown in the figure are examples for a single pixel, 3 base classes and 1 novel class for simplicity.
    Dashed boxes represent the frozen modules.}
    \label{fig:framework}
\end{figure*}

\section{Related Work}
\subsection{Few-shot Learning}
Few-Shot Learning (\textit{FSL}) aims at learning a model that can recognize novel classes with limited samples. 
There are two main streams for existing \textit{FSL} methods: optimization based \cite{cai2018memory,finn2017model,rusu2018meta} and metric-learning based \cite{snell2017prototypical,sung2018learning,vinyals2016matching,zhang2020deepemd,qi2018low}. 
Optimization based methods learn a model with good initialization that can adapt to novel classes fast.
For example, MAML \cite{finn2017model} proposed the model-agnostic meta-learning that trains a model on a variety of learning tasks for a fast adaptation on unseen classes.
Each task in MAML contains a small amount of training data and the model is trained on it for a handful of steps so that the model is generalized for a new task.
\cite{cai2018memory} proposed to combine a memory bank and meta-learning the network's parameters for unlabeled images for one-shot learning.
Typically, a contextual learner with sequential inputs from the memory bank slots is able to predict the parameters for unlabeled data.
Further, LEO \cite{rusu2018meta} focused on the unbalanced problem: high-dimensional parameter spaces \textit{vs.} extreme low-data regimes, by introducing latent embedding optimization.
In contrast, the purpose of metric-learning based methods is to learn a good embedding function which can generate a robust prototype for each novel class.
For instance, the popular Prototypical Networks \cite{snell2017prototypical} meta-learned a prototype for each novel class by averaging the latent embeddings in the support set.
The predicted class of a sample is determined by the closest prototype to its output feature.
The closeness is computed by Euclidean distance.
Instead of using existing distance metrics, \cite{sung2018learning} proposed to learn to learn a deep distance metric that compares the similarity between support and query samples.

Generalized Few-Shot Learning (\textit{GFSL}) \cite{hariharan2017low} extends \textit{FSL} to simultaneously recognize both base and novel classes.
Generally, two steps are needed for a \textit{GFSL} model: many-shot learning on base classes and few-shot learning on novel classes.
\cite{gidaris2018dynamic} introduced an attention based few-shot classification weight generator to generate a classification weight vector for novel class, which is then combined with the trained base class classifier weights for \textit{GFSL}.
ACASTLE \cite{ye2021learning} introduced a neural dictionary based Transformer to adjust the combined classifier weights. 
The common characteristic for these works is to do classifier parameter (weights) fusion.
In this work, we study Generalized Few-shot Semantic Segmentation (\textit{GFSS}), a similar but more challenging task compared to \textit{GFSL}.
Instead of fusing the base and novel classifiers at the  parameter-level, we propose to fuse at score level and design a Transformer-based score calibration method to avoid fusion bias.

\subsection{Few-shot Semantic Segmentation}
Few-shot Semantic Segmentation (\textit{FSS}) targets at the same few-shot challenge as \textit{FSL} but focuses on semantic segmentation task.
Existing \textit{FSS} methods generally build relationships between query images and support images in two ways: prototypical learning \cite{dong2018few,liu2020part,wang2019panet} and feature concatenation \cite{cao2020few,yang2020prototype,zhang2019canet,tian2020prior}.
PL \cite{dong2018few} introduced prototypical learning \cite{snell2017prototypical} into \textit{FSS}.
PANet \cite{wang2019panet} further proposed a prototype alignment regularization.
Recently, part-aware prototypes have gained much attention as investigated in PPNet \cite{liu2020part}. 
In contrast, feature concatenation based methods first combine the query features and support prototypes together and obtain support-conditioned query features as the input of the subsequent segmentation head.
For instance, PFENet \cite{tian2020prior} proposed feature enrichment network for the combination of the query features and support prototypes.
But the main limitation of \textit{FSS} methods is that the trained model is only able to segment few novel classes.
Apart from FSS, there are also other related semantic segmentation tasks.
Among them, Generalized Few-shot Semantic Segmentation (\textit{GFSS}) is the most challenging setting.
For better understanding, we show the differences among them in Table \ref{tab:setting_com}.

\begin{table}[ht]
    \centering
    \caption{Setting comparisons among Semantic Segmentation (SS), Few-shot Semantic Segmentation (FSS), Continual/Incremental Semantic Segmentation (CSS/ISS) and Generalized Few-shot Semantic Segmentation (GFSS).}
    \begin{tabular}{c|cccc}
    \hline
         Setting & SS & FSS & CSS/ISS & GFSS  \\
         \hline
         Training with base classes & $\surd$ & $\surd$ & $\surd$ & $\surd$ \\
         Training with novel classes & $\times$ & $\times$ & $\surd$ & $\times$ \\
         Sufficient samples per novel class & $\times$ & $\times$ & $\surd$ & $\times$ \\
         Test on base classes & $\surd$ & $\times$ & $\surd$ & $\surd$ \\
         Test on novel classes & $\times$ & $\surd$ & $\surd$ & $\surd$ \\
         Using base classes during testing & $\times$ & $\times$ & $\times$ & $\times$ \\
         Using novel classes during testing & $\times$ & $\surd$ & $\times$ & $\surd$ \\
    \hline
    \end{tabular}
    \label{tab:setting_com}
\end{table}

\textit{GFSS} \cite{liu2020dynamic} generalizes \textit{FSS} to segment both base and novel classes.
DENet \cite{liu2020dynamic} proposed a dynamic extension network which used a learning strategy to avoid bias towards base classes. 
CAPL \cite{tian2020generalized} proposed to combine base-class trained classifier's weights with novel ones obtained by averaging novel features.
However, their learned models are found to be poor on base classes \cite{liu2020dynamic} or novel classes \cite{tian2020generalized} (Table \ref{tab:pascal_main}). In this work, we propose a novel framework for \textit{GFSS} based on prediction score fusion, along with a Transformer based prediction calibration module, maintaining the good performance on both base and novel classes.

\subsection{Prediction Calibration}
Prediction calibration is a common technique used in open-set recognition. 
It relies on a threshold to classify the novel classes, i.e., its prediction score from base classifier is lower than the threshold \cite{hendrycks2016baseline}. 
However, deep models are usually overconfident to their inputs \cite{guo2017calibration,hein2019relu}, regardless of base or novel classes. Therefore, threshold-based calibration methods are typically poor at detecting novel classes. 
Different from the goal of prediction calibration for open-set recognition, we propose a prediction calibration network for \textit{GFSS} using a novel cross-covariance based Transformer.

\subsection{Transformers}
Recently, inspired by the great success of self-attention in NLP \cite{vaswani2017attention,devlin2019bert}, Transformer-based architectures have been explored in vision tasks, such as object detection \cite{hu2018relation,carion2020end}, semantic segmentation \cite{zheng2021rethinking}, and image classification \cite{dosovitskiy2020image,wu2020visual}. 
Our calibration Transformer is inspired by the recent  cross-covariance based vision Transformer~\cite{el2021xcit} with a shared goal for efficiency but different applications and formulations.

\section{Methodology}

\subsection{Problem setup} The ultimate goal of \textit{GFSS} is to learn a model that can segment images of both base and novel classes. 
A \textit{GFSS} model needs to be trained in three stages. 1) \emph{Base classifier training}: a base classifier is trained on the large amount of labeled data $D_{base}=\{x, y\}_{i}^{}$ of $N_{base}$ base classes ($C_{base}$).
2) \emph{Novel classifier training}: a novel classifier will be trained on a support set (1-5 shots per class) from ``fake'' novel\footnote{The ``fake'' novel classes are sampled from $C_{base}$ for episodic training.}/novel classes ($C_{novel}| C_{base} \cap C_{novel} = \emptyset$) during training/testing.
3) \emph{Base + novel class fusion}: the trained base and novel classifiers are fused and  tested on a test set $D_{test}$ including images of both base and novel classes.
Note that samples from base and novel classes do not co-exist in any stages.

\subsection{Overview} Let us denote the base and the novel models as $G_{base} \circ F_{base}$ and $G_{novel} \circ F_{novel}$ respectively, each comprising a feature extractor and a classifier. We assume the base and novel classifiers share the same feature extractor, i.e., $F_{base}=F_{novel}=F$.
Note that $F$ extracts multi-level fused features from Pyramid Pooling Module (PPM) of PSPNet \cite{zhao2017pyramid} (our backbone model).
Besides the main task models, we also have a  Normalized Score Fusion (NSF) module and a Transformer-based calibration module denoted as $\mathcal{T}$, which is episodically trained on $D_{base}$. 
The overall pipeline of our \textit{PCN} model is shown in Figure~\ref{fig:framework}.

\subsection{Base Classifier Training}
We first train our model $G_{base} \circ F_{base}$ on the base dataset $D_{base}$. The feature extractor $F$ and classifier $G_{base}$ are optimized using the cross entropy loss as follows:
\begin{equation}
\underset{F, G_{base}}{\arg\min}\quad \frac{1}{|D_{base}|}\sum_{x, y\sim D_{base}} \frac{1}{H\times W}\sum_{i=1}^{H\times W} \ell_{ce}(y_{i}, \hat{y}_i),  \\
    \label{base_clf}
\end{equation}
where $\hat{y}=S(G_{base}\circ F(x))$ and $(H,W)$ is the shape of input $x$ and mask $y$. $S$ is the softmax function. After training, both the feature extractor and the base classifier are frozen in the next stages.

\subsection{Support-set based Novel Classifier Training}
With a pre-trained feature extractor $F$, we train a novel classifier $G_{novel}$ using a support set $D_{novel}$ with just a few samples per novel class. The optimization objective is as follows,
\begin{equation}
\underset{G_{novel}}{\arg\min}\quad \frac{1}{|D_{novel}|}\sum_{x, y\sim D_{novel}} \frac{1}{H\times W}\sum_{i=1}^{H\times W} \ell_{ce}(y_{i}, \hat{y}_i),
    \label{novel_clf}
\end{equation}
where $\hat{y}=S(G_{novel}\circ F(x))$.
Once trained, the novel classifier needs to be fused with the base classifier to segment images of both base and novel classes.

\subsection{Prediction Calibration}
\label{stage2}
Existing approach to \textit{GFSS} is based on parameter fusion. The prediction is formulated as
$$
\hat{y} = S((G_{base}\cup G_{novel}) \circ F(x)).
$$
However, with much more base classes and training samples, the fused classifier will be biased towards the base classes. Instead of parameter-level fusion, we adopt a score-level fusion approach, i.e., for each pixel, the two classifiers generate two sets of scores separately. These two sets of scores are then fused with calibration to avoid bias towards either the base or novel classes. To that end,  two modules are applied sequentially: Normalized Score Fusion (NSF) and Transformer-based prediction calibration.

\paragraph{Normalized Score Fusion} 
With NSF, we normalize the predicted scores of two classifiers separately to promote the saliency of the novel classes as 
\begin{equation}
\label{eq:nsf}
    \hat{y}_{nsf} = S(G_{base} \circ F(x)) \cup S(G_{novel} \circ F(x)).
\end{equation}
Empirically, we find that this simple technique mitigates the prediction bias towards base classes and already yields the state of the art performance. However, the prediction tends to be overconfident on novel classes as shown in  Figure~\ref{fig:visacm}.

\paragraph{Transformer-based prediction calibration} In order to further improve the prediction calibration, we propose a feature-score cross-covariance based Transformer $\mathcal{T}$ for prediction calibration. Concretely, as per standard, our Transformer also has three projection heads: Query head $\Omega: \mathbb{R}^{c\times hw} \rightarrow \mathbb{R}^{c\times d}$, Key head $\Gamma: \mathbb{R}^{m\times hw} \rightarrow \mathbb{R}^{m\times d}$ and Value head $\Lambda: \mathbb{R}^{m\times hw} \rightarrow \mathbb{R}^{m\times d}$ ($c$: the class number, $m$: feature dimension, $d$: projected dimension, $h,w$: spatial dimension). Different from the conventional design, which treats each pixel/patch embedding as a token, we instead treat each score and feature channel as our token and compute the cross covariance between them as
\begin{equation}
\label{eq:cov}
  \sigma  = \Omega(\hat{y}) \cdot (\Gamma(F(x)))^T, \sigma \in \mathbb{R}^{c \times m}.
\end{equation}
This significantly reduces the Transformer computational cost and makes it tractable for our pixel-level global attention task. Then, the value embedding is transformed as
\begin{equation}
\label{eq:pc1}
  v = \frac{S(\sigma)}{\sqrt{d}} \cdot \Lambda(F(x)).
\end{equation}
After that, a calibration head $\Delta: \mathbb{R}^{c\times d} \rightarrow \mathbb{R}^{c\times hw}$ is followed to predict the prediction offset to calibrate the vanilla model predictions as,
\begin{equation}
\label{eq:pc2}
\begin{aligned}
  \hat{y}_{\Delta} & = \Delta(v) \\
  \hat{y}_{calib} & = \hat{y}_{nsf} + \hat{y}_{\Delta}.
\end{aligned}
\end{equation}

\paragraph{Episodic Training} 
\label{ep_t}
As we have no access to the unseen novel classes during training, to simulate the testing scenarios, in each episode we randomly sample $N_{fn}$ ``fake'' novel classes ($C_{fn}$) from base dataset $D_{base}$ as per \cite{gidaris2018dynamic,ye2021learning}. For each ``fake'' novel class, we sample $K$ support images as well as $Q$ query images. We sample the same number of query images from the remaining $N_{rb}$ base classes ($C_{rb}$) to balance the training ($N_{fn} + N_{rb} = N_{base}$, and $C_{fn} \cap C_{rb} = \emptyset$ ). The sampled support and query sets can be denoted as:
\begin{equation}
    \begin{aligned}
        & \mathcal{S} = \{(x_i, y_i)~|~ y_{i} \in C_{fn}\}^{N_{fn}K}_{i=1} \\
        &\mathcal{Q} = \{(x_{i}, y_{i})~|~y_{i}\in C_{fn}\}^{N_{fn}Q}_{i=1} \ \cup \\ 
        &\ \ \ \ \ \ \ \{(x_{i}, y_{i})~|~y_{i} \in C_{rb}\}^{N_{fn}Q}_{i=1}.
    \end{aligned}
\end{equation}

In each episode, we first activate the classifier weights $G_{rb}$ for the real base classes from the pre-trained base classifier $G_{base}$. Meanwhile, we use the support set of the novel classes to train a ``fake'' novel classifier as 
\begin{equation}
\label{eq:fn_loss}
\begin{aligned}
G_{fn} &= \underset{G_{fn}}{\arg\min}\quad \mathcal{L}_{ce}(y_{i}, \mathcal{S}).
\end{aligned}
\end{equation}

Then, we sample the ``fake'' test data from the query set $\mathcal{Q}$ and optimize our proposed calibration module. 
Given its NSF predictions $\hat{y}_{fnsf}$ of the real base and ``fake'' novel classifiers $G_{rb}$ and $G_{fn}$, we can get the calibrated predictions $\hat{y}_{fcalib}$ through our Transformer-based calibration module as $\hat{y}_{fcalib} = \mathcal{T}_{(\Omega, \Gamma, \Lambda, \Delta)}(F(x), \hat{y}_{fnsf})$. Then, we can compute the cross entropy loss per pixel and optimize the calibration module as 
\begin{equation}
\label{eq:pcn_loss}
\begin{aligned}
\underset{\Omega, \Gamma, \Lambda, \Delta}{\arg\min}\quad \frac{1}{|\mathcal{Q}|}\sum_{x, y\sim \mathcal{Q}} \frac{1}{H\times W}\sum_{i=1}^{H\times W} \ell_{ce}(y_{i}, \hat{y}_{fcalib}).
\end{aligned}
\end{equation}

The overall training algorithm could be found in Algorithm~\ref{alg:meta_pcn}.

\RestyleAlgo{ruled}

\begin{algorithm}[t]
\caption{Episodic Training for \textit{PCN}}\label{alg:meta_pcn}
\KwData{Support set $\mathcal{S}$ and query set $\mathcal{Q}$ from $C_{base}$;}
Initialize Transformer module $\mathcal{T}$ with ${\theta}_{t}$\;
Set learning rate $\alpha_t$ for $\mathcal{T}$\;
\For{each iteration}{
Activate $G_{rb}$ with $G_{base}$\;
Obtain $\{(x_i^S, y_i^S) | y_i^S \in C_{fn}\}$ from $\mathcal{S}$\;
Initialize a novel classifier $G_{fn}$ with ${\theta}_{n}$\;
Set learning rate $\alpha_n$ for $G_{fn}$\;
\While{train a novel classifier}{
    Predict $\hat{y}_i^S$\ by $G_{fn} \circ F(x_i^S)$\;
    Compute the loss $\mathcal{L}_n$ via Ep. \ref{eq:fn_loss}\;
    Update $G_{fn}$ by ${\theta}_{n} \leftarrow {\theta}_{n} - \alpha_n {\bigtriangledown}_{{\theta}_{n}} \mathcal{L}_{n}$\;
}
Obtain $\{(x^Q, y^Q) | y^Q \in C_{fn} \cup C_{rb}\}$ from $\mathcal{Q}$\;
Predict $\hat{y}_{rb}^Q$ by $G_{rb} \circ F(x^Q)$\; 
Predict $\hat{y}_{fn}^Q$ by $G_{fn} \circ F(x^Q)$\;
Normalized Score Fusion via Ep. \ref{eq:nsf} $\rightarrow$ $\hat{y}_{fnsf}^Q$\; 
Compute the cross covariance $\sigma$ via Ep. \ref{eq:cov}\;
Value transformation via Ep. \ref{eq:pc1}\;
Compute prediction offset via Ep. \ref{eq:pc2} \;
Obtain calibrated prediction score: $\hat{y}_{fcalib}^Q$\;
Compute the loss $\mathcal{L}_t$ via Ep. \ref{eq:pcn_loss}\;
Update $\mathcal{T}$ by $\theta_t \leftarrow \theta_t - \alpha_t {\bigtriangledown}_{{\theta}_{t}} \mathcal{L}_t$\;
}
\end{algorithm}

\subsection{Inference}
Given a pre-trained base model $G_{base} \circ F$, a support set trained novel model $G_{novel} \circ F$ and a learned calibration module $\mathcal{T}$, our model can now segment any unseen data from $D_{test}$ with both the base and novel classes. The overall inference formulation is 
\begin{equation}
    \begin{aligned}
        \hat{y} = \mathcal{T}(F(x), \hat{y}_{nsf}), x\sim D_{test},
    \end{aligned}
\end{equation}
where $
\hat{y}_{nsf}$ is computed using Eq.~\ref{eq:nsf}.

\section{Experiments}
\subsection{Datasets} We conduct experiments on two widely used datasets in \textit{FSS}, i.e., PASCAL-$5^{i}$ and COCO-$20^{i}$. 
PASCAL-$5^{i}$ is the PASCAL-VOC 2012 dataset \cite{everingham2010pascal} with augmented annotations from \cite{hariharan2014simultaneous}, and has 5,953/1,449 images in train/val set and 20 classes in each set. 
As per \cite{shaban2017one}, 4 splits $i\in \{0,1,2,3\}$ are separated with 5 classes in each one for cross-validation.
In each experiment, we use 3 splits as the base class training set without having access to another split from novel classes. 
During testing on the val set, the classes unseen in training construct the novel dataset while seen classes are used as the base query.
In total, four experiments are conducted on this dataset. 
COCO-$20^{i}$ is a relatively larger dataset, with 82,081/40,137 images in 80 classes in the train/val set respectively. 
Following \cite{nguyen2019feature}, it was split into four folds. 
Each fold has 20 classes. 
Experiments are conducted in a cross-validation manner as PASCAL-$5^{i}$.

\begin{table*}[ht]
    \centering
    \caption{Generalized few-shot semantic segmentation results on PASCAL-$5^{i}$. The results of Base Classifier show the upper bound performance on base classes. NSF: normalized score fusion only.}
    \begin{adjustbox}{max width=\textwidth}
    \begin{tabular}{l|l|c|cccccccc|cccc}
    \hline
         \multirow{2}{*}{Backbone} & \multirow{2}{*}{Method} & \multirow{2}{*}{Shot} & \multicolumn{2}{c}{s-0} &  \multicolumn{2}{c}{s-1} &  \multicolumn{2}{c}{s-2} &  \multicolumn{2}{c}{s-3} & \multicolumn{4}{c}{Average}  \\
         \cline{4-15}
         & & & Base & Novel & Base & Novel & Base & Novel & Base & Novel & Base & Novel & mIoU & $H_{mean}$ \\
         \hline
         \multirow{11}{*}{ResNet-50} & Base Classifier & - & 63.15 & - & 64.17 & - & 63.87 & - & 65.47 & - & 64.17 & - & - & - \\
         \cline{3-15}
         & DENet (MM20) & \multirow{6}{*}{1} & 29.91 & \textbf{56.66} & 18.20 & \textbf{65.68} & 20.75 & \textbf{60.48} & 22.45 & \textbf{47.02} & 22.83 & \textbf{57.46} & 31.49 & 32.67 \\
         & PFENet (PAMI20) & & 58.24 & 13.00 & 59.14 & 9.99 & 54.87 & 6.08 & 58.61 & 5.13 & 57.72 & 8.55 & 45.42 & 14.89 \\
         & ACASTLE (IJCV21) & & \textbf{59.46} & 34.82 & 55.12 & 37.64 & 59.69 & 40.45 & 61.09 & 32.08 & 58.84 & 36.25 & 53.19 & 44.86 \\
         & CAPL (CVPR22) & & - & - & - & - & - & - & - & - & 65.48 & 18.85 & 54.38 & 29.16 \\
         & NSF (Ours) & & 29.38 & 51.86 & 36.61 & 55.41 & 44.73 & 50.63 & 42.30 & 36.93 & 38.26 & 48.71 & 40.87 & 42.85 \\
         & PCN (Ours) & & 59.43 & 47.98 & \textbf{61.31} & 51.27 & \textbf{61.44} & 51.25 & \textbf{62.93} & 41.34 & \textbf{61.28} & 47.96 & \textbf{57.95} & \textbf{53.81} \\
         \cline{2-15}
         & DENet (MM20) & \multirow{6}{*}{5} & 30.15 & 58.04 & 18.58 & \textbf{66.10} & 21.21 & 61.00 & 22.84 & 47.73 & 23.20 & 58.22 & 31.95 & 33.17 \\
         & PFENet (PAMI20) & & 57.81 & 13.38 & 58.96 & 9.80 & 54.72 & 6.17 & 58.68 & 5.01 & 57.54 & 8.59 & 45.30 & 14.95 \\
         & ACASTLE (IJCV21) & & \textbf{60.56} & 45.64 & \textbf{62.87} & 53.87 & \textbf{62.91} & 42.41 & \textbf{63.59} & 42.68 & \textbf{62.48} & 46.15 & 58.40 & 53.09 \\
         & CAPL (CVPR22) & & - & - & - & - & - & - & - & - & 66.14 & 22.41 & 55.72 & 33.48 \\
         & NSF (Ours) & & 29.90 & \textbf{59.07} & 44.99 & 64.01 & 49.60 & \textbf{62.74} & 48.25 & 48.36 & 43.19 & \textbf{58.55} & 47.03 & 49.71 \\
         & PCN (Ours) & & 59.05 & 53.03 & 60.59 & 58.03 & 60.60 & 61.67 & 63.15 & \textbf{51.60} & 60.85 & 56.08 & \textbf{59.66} & \textbf{58.37} \\
         \hline
         \multirow{9}{*}{ResNet-101} & Base Classifier & - & 67.42 & - & 66.08 & - & 64.93 & - & 67.70 & - & 66.53 & - & - & - \\
         \cline{3-15}
         & PFENet (PAMI20) & \multirow{4}{*}{1} & 58.64 & 20.42 & 59.72 & 6.00 & 56.85 & 12.62 & 55.06 & 13.31 & 57.57 & 13.09 & 46.45 & 21.33 \\
         & ACASTLE (IJCV21) & & \textbf{62.30} & 39.61 & 56.60 & 38.12 & 59.43 & 32.17 & 55.84 & 30.34 & 58.54 & 35.06 & 52.67 & 43.86 \\
         & NSF (Ours) & & 27.65 & \textbf{52.64} & 34.42 & \textbf{61.93} & 37.75 & \textbf{58.72} & 41.63 & \textbf{41.56} & 35.36 & \textbf{53.71} & 39.95 & 42.65 \\
         & PCN (Ours) & & 60.30 & 51.38 & \textbf{60.04} & 44.03 & \textbf{61.16} & 39.91 & \textbf{61.29} & 33.09 & \textbf{60.70} & 42.10 & \textbf{56.05} & \textbf{49.72} \\
         \cline{2-15}
         & PFENet (PAMI20) & \multirow{4}{*}{5} & 58.56 & 20.65 & 59.70 & 6.48 & 57.02 & 12.20 & 54.71 & 13.17 & 57.50 & 13.13 & 46.40 & 21.37 \\
         & ACASTLE (IJCV21) & & \textbf{62.98} & 51.75 & \textbf{63.26} & 47.61 & \textbf{64.04} & 41.07 & \textbf{62.82} & 36.28 & \textbf{63.28} & 44.18 & \textbf{58.50} & 52.03 \\
         & NSF (Ours) & & 27.41 & \textbf{59.97} & 41.54 & \textbf{67.39} & 43.44 & \textbf{65.69} & 47.96 & \textbf{50.47} & 40.09 & \textbf{60.88} & 45.29 & 48.34 \\
         & PCN (Ours) & & 59.10 & 56.35 & 59.18 & 49.21 & 60.25 & 48.41 & 60.88 & 43.47 & 59.85 & 49.36 & 57.23 & \textbf{54.10} \\
         \hline
    \end{tabular}
    \end{adjustbox}
    \label{tab:pascal_main}
\end{table*}

\begin{table*}[ht]
    \centering
    \caption{Generalized few-shot semantic segmentation results on COCO-$20^{i}$. The results of Base Classifier show the upper bound performance on base classes. NSF: normalized score fusion only.}
    \begin{adjustbox}{max width=\textwidth}
    \begin{tabular}{l|l|c|cccccccc|cccc}
    \hline 
         \multirow{2}{*}{Backbone} & \multirow{2}{*}{Method} & \multirow{2}{*}{Shot} & \multicolumn{2}{c}{s-0} &  \multicolumn{2}{c}{s-1} &  \multicolumn{2}{c}{s-2} &  \multicolumn{2}{c}{s-3} & \multicolumn{4}{c}{Average}  \\
         \cline{4-15}
         & & & Base & Novel & Base & Novel & Base & Novel & Base & Novel & Base & Novel & mIoU & $H_{mean}$ \\
         \hline
         \multirow{11}{*}{ResNet-50} & Base Classifier & - & 44.29 & - &	34.14 & - & 36.29 & - &	37.47 & - & 38.05 & - & - & - \\
         \cline{3-15}
         & DENet (MM20) & \multirow{6}{*}{1} & 11.79 & \textbf{37.37} & 10.42 & \textbf{38.23} & 12.82 & \textbf{33.68} & 11.33 & \textbf{34.67} & 11.59 & \textbf{35.99} & 17.69 & 17.53 \\
         & PFENet (PAMI20) & & \textbf{39.88} & 4.22 & 28.40 & 6.17 & 31.23 & 3.88 & 31.75 & 2.13 & 32.82 & 4.10 & 25.64 & 7.29 \\
         & ACASTLE (IJCV21) & & 23.42 & 25.57 & 21.93 & 21.64 & 25.17 & 20.40 & 25.90 & 18.43 & 24.11 & 21.51 & 23.46 & 22.73 \\
         & CAPL (CVPR22) & & - & - & - & - & - & - & - & - & 44.61 & 7.05 & \textbf{35.46} & 12.18 \\
         & NSF (Ours) & & 30.99 & 32.26 & 22.85 & 32.12 & 25.47 & 29.32 & 27.59 & 23.69 & 26.73 & 29.35 & 27.38 & 27.97 \\
         & PCN (Ours) & & 39.83 & 31.93 & \textbf{33.60} & 28.63 & \textbf{35.33} & 27.53 & \textbf{36.51} & 29.39 & \textbf{36.32} & 29.37 & 34.58 & \textbf{32.48} \\
         \cline{2-15}
         & DENet (MM20) & \multirow{6}{*}{5} & 12.30 & 38.78 & 10.81 & 39.73 & 13.13 & 34.77 & 11.43 & 35.10 & 11.92 & 37.10 & 18.21 & 18.04 \\
         & PFENet (PAMI20) & & 39.83 & 4.38 & 28.18 & 6.00 & 31.20 & 3.77 & 31.49 & 2.23 & 32.68 & 4.10 & 25.53 & 7.28 \\
         & ACASTLE (IJCV21) & & \textbf{42.03} & 23.78 & 22.32 & 26.11 & \textbf{35.22} & 17.52 & 35.60 & 23.63 & 33.79 & 22.76 & 31.03 & 27.20 \\
         & CAPL (CVPR22) & & - & - & - & - & - & - & - & - &  45.24 & 11.05 & \textbf{36.80} & 17.76 \\
         & NSF (Ours) & & 33.44 & \textbf{41.78} & 24.22 & \textbf{39.85} & 26.43 & \textbf{36.12} & 29.61 &	36.03 &	28.43 &	\textbf{38.45} & 30.93 & 32.68 \\
         & PCN (Ours) & & 40.16 & 40.82 & \textbf{33.42} & 35.76 & 34.86 & 33.04 & \textbf{36.31} & \textbf{38.36} & \textbf{36.19} & 37.00 & 36.39 & \textbf{36.59} \\
         \hline
         \multirow{9}{*}{ResNet-101} & Base Classifier & - & 37.49 & - & 41.72 & - & 37.28 & - &	41.85 & - &	39.59 & - & - & - \\
         \cline{3-15}
         & PFENet (PAMI20) & \multirow{4}{*}{1} & 34.44 & 2.81 & 38.35 & 2.55 & 34.83 & 5.07 & 38.38 & 3.90 & 36.50 & 3.58 & 28.27 & 6.52\\
         & ACASTLE (IJCV21) & & 21.70 & 25.90 & 28.53 & 23.57 & 20.21 & 18.75 & 29.83 & 21.99 & 25.07 & 22.55 & 24.44 & 23.74 \\
         & NSF (Ours) & & 25.08 & \textbf{32.66} & 31.74 & \textbf{33.33} & 27.70 & \textbf{30.50} & 32.49 & \textbf{30.32} & 29.25 & \textbf{31.70} & 29.87 & 30.43 \\
         & PCN (Ours) & & \textbf{35.26} & 27.54 & \textbf{39.08} & 30.32 & \textbf{36.07} & 24.66 & \textbf{39.46} & 26.37 & \textbf{37.47} & 27.22 & \textbf{34.91} & \textbf{31.53} \\
         \cline{2-15}
         & PFENet (PAMI20) & \multirow{4}{*}{5} & 34.52 & 2.53 & 38.32 & 2.76 & 34.69 & 6.09 & 38.72 & 3.64 & 36.56 & 3.76 & 28.36 & 6.81 \\
         & ACASTLE (IJCV21) & & \textbf{35.33} & 20.72 & \textbf{41.22} & 23.75 & 34.01 & 22.77 & \textbf{40.97} & 19.34 & \textbf{37.88} & 21.65 & 33.82 & 27.55 \\
         & NSF (Ours) & & 27.18 & \textbf{40.74} & 34.24 & \textbf{41.94} & 29.95 & \textbf{37.16} & 34.69 & \textbf{39.23} & 31.52 & \textbf{39.77} & 33.58 & 35.16 \\
         & PCN (Ours) & & 35.09 & 35.84 & 40.05 & 38.11 & \textbf{35.75} & 31.01 & 40.50 & 33.73 & 37.85 & 34.67 & \textbf{37.05} & \textbf{36.19} \\
         \hline
    \end{tabular}
    \end{adjustbox}
    \label{tab:coco_main}
\end{table*}

\subsection{Implementation Details} In stage one, we train the feature extraction network $F$ and the base classifier $G_{base}$ on the base dataset. We adopt PSPNet \cite{zhao2017pyramid} as our feature extraction backbone with two versions, i.e., ResNet-50 and ResNet-101.
The training epochs are 100 and 20 for PASCAL-$5^{i}$ and COCO-$20^{i}$, respectively. 
The batch size is 12 and the input image size is 417. The latent feature map's spatial dimension $h=w=53$. We use momentum SGD to train our model. 
The initial learning rate is $2.5e-3$, decayed by a cosine learning rate scheduler during training. 

In the second episodic-training stage, in each episode we first use SGD optimizer to train a novel class classifier $G_{novel}$ (a 3 $\times$ 3 Conv layer $\rightarrow$ a ReLU activation $\rightarrow$ a 1 $\times$ 1 Conv layer) on a support set with 50 iterations. 
The learning rate is  $1e-1$ and fixed. 
Weighted cross-entropy is used to stabilize the training. 
After that, we meta-learn our \textit{PCN} within the same episode.
We train our models for 10,000/30,000 iterations with batch size 8 on PASCAL-$5^{i}$/COCO-$20^{i}$. The initial learning rate for \textit{PCN} is set to $1e-2$, which is also decayed by a cosine learning rate scheduler. 
We set $N_{fn}=1$ and $Q=1$ during training.

\subsection{Evaluation Metrics}

According to generalized few shot recognition setting \cite{gidaris2018dynamic,ye2021learning}, we adopt harmonic mean Intersection over Union (mIoU) to evaluate our model. 
Specifically, in each testing task, we sample $N_{base}$\footnote{The total number of base classes, e.g., 15/60 for PASCAL-$5^{i}$/COCO-$20^{i}$.} pairs of query images. 
Each pair has two images, one from novel class and the other one from the $j^{th} \in \{ 1,\dots, N_{base}\}$ base class. 
We separately compute the mIoU over the base classes ($mIoU_{base}$) and novel classes ($mIoU_{novel}$) and then compute their harmonic mean ($H_{mean}$) by:
\begin{equation}
    H_{mean} = \frac{2 \cdot mIoU_{base} \cdot mIoU_{novel}}{mIoU_{base} + mIoU_{novel}}.
\end{equation}
We also report mIoU over all classes ($mIoU$).
On each testing split, we evaluate our model with 1,000 tasks for both PASCAL-$5^{i}$ and COCO-$20^{i}$. 

\subsection{Compared Methods}

We compare our method with DENet \cite{liu2020dynamic}, CAPL \cite{tian2020generalized}, PFENet \cite{tian2020prior} and ACASTLE \cite{ye2021learning}. 
DENet and CAPL are two state-of-the-art \textit{GFSS} methods. PFENet is a state-of-the-art \textit{FSS} model which is naturally extendable to \textit{GFSS}. 
ACASTLE is a state-of-the-art generalized few-shot recognition model, which can also be re-purposed for \textit{GFSS}. 
All results are produced based on their officially released codes except CAPL \cite{tian2020generalized} that does not release code (citing results from their paper). 
In addition, a baseline using normalized score fusion only is also compared.

\subsection{PASCAL-$5^{i}$ Results}
\begin{figure*}[ht]
    \centering
    \includegraphics[width=0.95\textwidth]{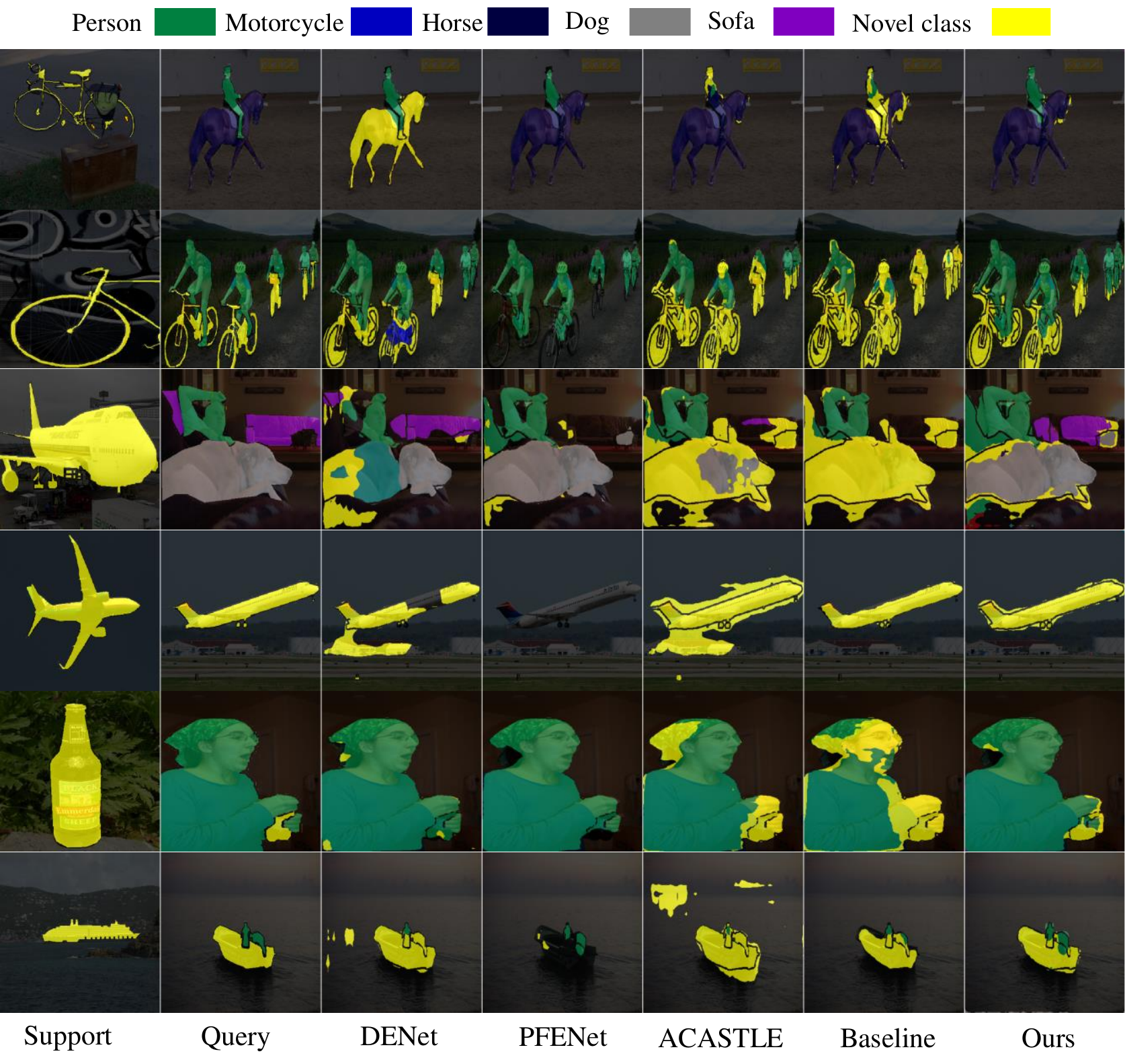}
    \caption{Qualitative results comparing our method against DENet, PFENet, ACASTLE and Normalized Score Fusion only (NSF). Setting: PASCAL-5$^{0}$ in the 1-shot case with ResNet-50 backbone for feature extraction. Each row contains an inference task with one novel class to be added to the segmentation.}
    \label{fig:vis_com}
\end{figure*}

The results on PASCAL-$5^{i}$ for 1-shot/5-shot case with ResNet-50/ResNet-101 are shown in Table \ref{tab:pascal_main}.
Overall, our method achieves a new state-of-the-art performance for all cases, with significant improvements over other compared methods.
Specifically, our method surpasses the best competitor by 8.95\%/5.86\% in 1-shot case and 5.28\%/2.07\% in 5-shot case, with ResNet-50/ResNet-101 respectively.
A number of other observations can be made.
First, the state-of-the-art \textit{FSS} method PFENet \cite{tian2020prior} has poor performance on novel classes when used for \textit{GFSS}.
The potential reason is that PFENet is learned on base classes and tested on novel classes only for \textit{FSS}. 
When the base and novel classes are tested together for \textit{GFSS}, PFENet is more confident on the base classes it was trained on and less confident on the unseen novel classes, yielding poor performance on novel classes. 
Second, other baselines including our NSF usually have an unbalanced performance. 
For example, DENet \cite{liu2020dynamic} generally has a much higher mIoU on novel classes while performs poorly on base classes. 
In contrast, our \textit{PCN} has a much more stable and balanced performance on both base and novel classes in most cases.
Third, except DENet and PFENet, all other methods benefit from more examples in the support set, cf. 1 shot vs 5 shot, on novel classes .
At last, our \textit{PCN} also wins in most cases with the mIoU over all classes.
Some qualitative results can be found in Figure \ref{fig:vis_com}. Overall, they are consistent with the quantitative observations.
For example, PFENet perfectly segments all objects of base classes in the $1^{st}, 5^{th}$ and $6^{th}$ rows while mis-segments the novel class \textit{Bike/Airplane/Bottle/Boat} in the $2^{nd}/4^{th}/5^{th}/6^{th}$ row.
In general, we can see our method can segment both base and novel classes more accurately in different scenarios.

\begin{table}[ht]
    \centering
    \caption{Comparisons among score fusion, Normalized Parameter Fusion (NPF) and our NSF. Dataset: PASCAL-5$^{0}$. Backbone: ResNet-50.}
    \begin{tabular}{l|c|ccccc}
    \hline
         \multirow{2}{*}{Method} & \multirow{2}{*}{Shot} & \multicolumn{4}{c}{s-0} \\
         \cline{3-6}
          & & Base & Novel & mIoU & $H_{mean}$ \\
         \hline
         Score Fusion &\multirow{3}{*}{1}  & 59.37 & 16.74 & 48.71 & 26.12 \\
         NPF & & \textbf{62.81} & 16.00 & \textbf{51.11} & 25.51 \\
         NSF & & 29.38 & \textbf{51.86} & 35.00 & \textbf{37.51} \\
    \hline
    \end{tabular}
    \label{tab:score_fusion}
\end{table}

\begin{figure*}[ht]
    \centering
    \includegraphics[width=0.95\textwidth]{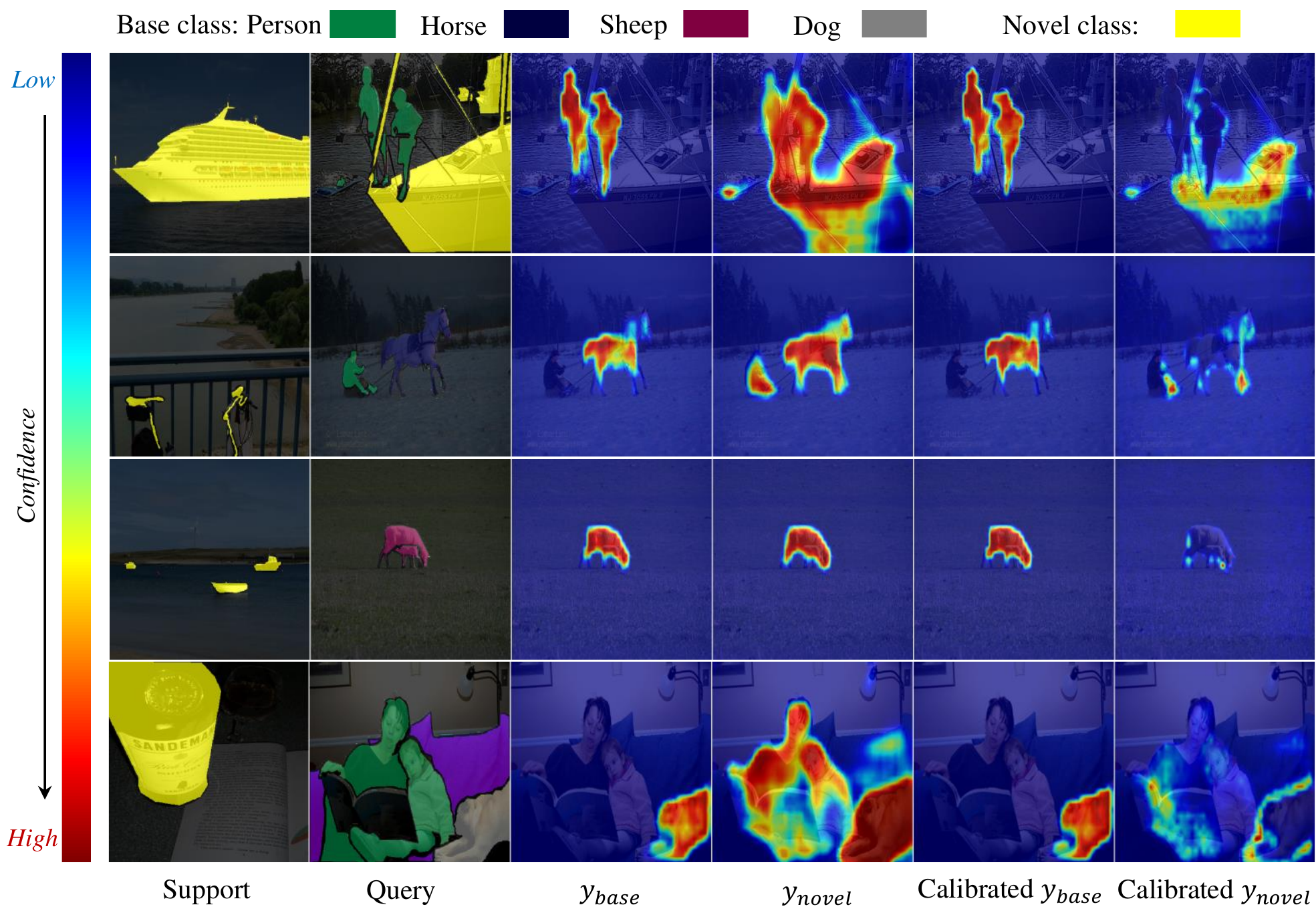}
    \caption{Visualization of the class prediction heatmap from base classifier and novel classifier before and after calibration. Note that for $y_{base}$ we only show one base class heatmap when multiple base classes exist. The main purpose of this figure is to show how the novel classifier's overconfidence on base classes is calibrated, which can be observed by comparing the $4^{th}$ and the last column.}
    \label{fig:atten_com}
\end{figure*}

\subsection{COCO-$20^{i}$ Results}
Table \ref{tab:coco_main} shows the comparative results on COCO-20$^{i}$.
Interestingly, our method is more superior over other baselines in this more challenging setup.
Specifically, our method achieves the overall best performance on base class mIoU, all class mIoU and $H_{mean}$.
Apart from the similar observations on PASCAL-$5^{i}$, there are some new findings.
(1) NSF only already outperforms all other competitors. 
This indicates that our proposed NSF is a promising strategy for mitigating the prediction bias.
(2) Our \textit{PCN} can further boost the performance by addressing novel classifier's overconfidence problem.
(3) We also notice that the performance gap is less significant with ResNet-101 as feature extraction backbone.
The possible reason is that the stronger backbone ResNet-101 might have better intrinsic novel class generalization ability, thus resulting in less improvements brought by our Transformer.

\subsection{Further Analysis}

\begin{figure}[t]
    \centering
    \includegraphics[width=0.48\textwidth]{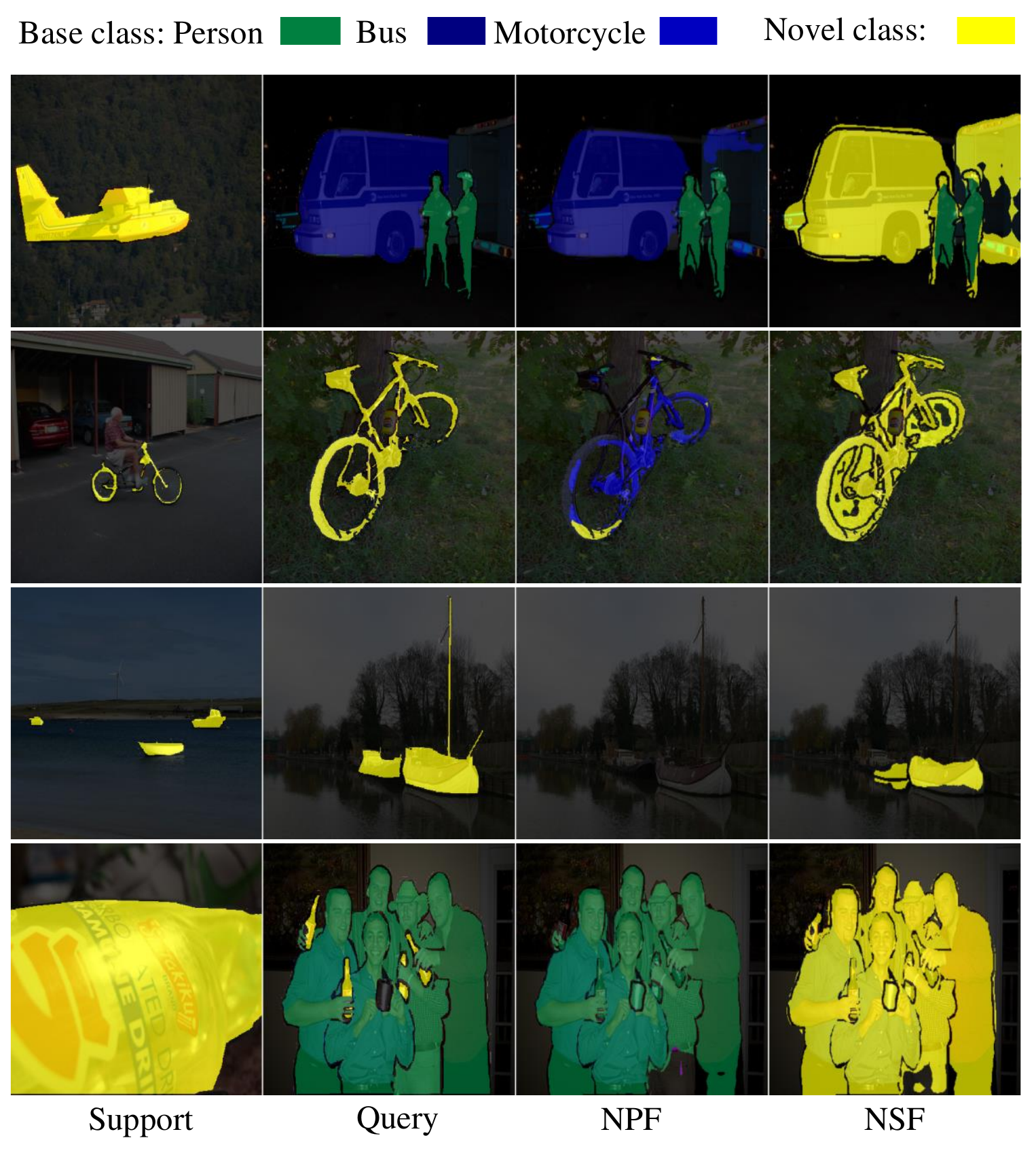}
    \caption{Visual comparisons between Normalized Parameter Fusion (NPF) and Normalized Score Fusion (NSF). From the results, we can see that NPF often mis-segments novel classes into base classes or does not detect them (in the $3^{rd}$ row) while NSF tends to segment base classes into novel ones.}
    \label{fig:joint_cls_pred}
\end{figure}

\paragraph{The Effectiveness of Normalized Score Fusion}
We also conduct experiments to analyze the effectiveness of our proposed normalized score fusion. From the results in Table~\ref{tab:score_fusion}, we can see that without normalization the naive score fusion can only achieve $26.12$ $H_{mean}$, showing its poor prediction calibration. 
We also experiment with an alternative Normalized Parameter Fusion (NPF), which normalizes the fused classifier parameters (weights) on feature channel before calculating the scores. 
However, we can see that this alternative performs as bad as the naive score fusion. 
Our normalized score fusion achieves the best among them with over $12.0$ $H_{mean}$ improvement. We visualize the results of NPF and NSF in Figure \ref{fig:joint_cls_pred}, which suggests that NPF performs poorly on novel classes while NSF are worse at base classes.

\paragraph{The Effectiveness of Our Transformer $\mathcal{T}$}
In this experiment, we first incorporate our proposed Transformer $\mathcal{T}$ with score fusion, NPF and our NSF. From the results in Table~\ref{tab:joint_cls}, we can see that it improves them noticeably with $\sim 18.0$, $\sim 17.0$ and $\sim 16.0$ $H_{mean}$ margins compared with Table~\ref{tab:score_fusion}. This demonstrates both the efficacy and generality of our proposed Transformer module.

Furthermore, we investigate the effectiveness of $\mathcal{T}$ on existing methods as shown in Table \ref{tab:previous}. Interestingly, the proposed $\mathcal{T}$ can always improve their performance on either base class or novel class so as to achieve the better overall mIoU and $H_{mIoU}$. For the method that has the largest unbalanced performance gap, i.e., PFENet, $\mathcal{T}$ boosts its performance considerably by 22.11\% on the harmonic mIoU.

\begin{table}[ht]
    \centering
    \caption{The effectiveness of our proposed Transformer. Dataset: PASCAL-5$^{0}$. Backbone: ResNet-50.}
    \begin{tabular}{l|c|ccccc}
    \hline
         \multirow{2}{*}{Method} & \multirow{2}{*}{Shot} & \multicolumn{4}{c}{s-0} \\
         \cline{3-6}
          & & Base & Novel & mIoU & $H_{mean}$ \\
         \hline
        Score Fusion + $\mathcal{T}$ & \multirow{3}{*}{1} & 59.31 & 35.68 & 53.40 & 44.55 \\
         NPF + $\mathcal{T}$ & & 46.85 & 39.21 & 44.94 & 42.69 \\
         NSF + $\mathcal{T}$ & & \textbf{59.43} & \textbf{47.98} & \textbf{56.57} & \textbf{53.09} \\
    \hline
    \end{tabular}
    \label{tab:joint_cls}
\end{table}

\begin{table}[ht]
    \centering
    \caption{Evaluation for previous methods equipped with the proposed Transformer $T$. Dataset: PASCAL-$5^{0}$. Backbone: ResNet-50. CAPL [29] did not release source code, so was not compared.}
    \begin{tabular}{l|c|ccccc}
    \hline
         \multirow{2}{*}{Method} & \multirow{2}{*}{Shot} & \multicolumn{4}{c}{s-0} \\
         \cline{3-6}
          & & Base & Novel & mIoU & $H_{mean}$ \\
         \hline
         DENet & \multirow{6}{*}{1} & 29.91 & \textbf{56.66} & 36.60 & 39.15  \\
         DENet + $T$ & & \textbf{39.54} & 52.98 & \textbf{42.90} & \textbf{45.28}  \\
         \cline{3-6}
         PFENet & & \textbf{58.24} & 13.00 & 46.93 & 21.26 \\
         PFENet + $T$ & & 51.35 & \textbf{37.44} & \textbf{48.01} & \textbf{43.37} \\
         \cline{3-6}
         ACASTLE & & 59.46 & 34.82 & 53.30 & 43.92 \\
         ACASTLE + $T$ & & \textbf{60.02} & \textbf{40.04} & \textbf{55.03} & \textbf{48.04}  \\
    \hline
    \end{tabular}
    \label{tab:previous}
\end{table}

\paragraph{Qualitative Results of Calibration}
The heatmaps of the class predictions from base classifier and novel classifier are visualized in Figure~\ref{fig:atten_com}. The middle two and rightmost two are generated from NSF and our full model respectively.
We can see that the novel classifier has high confidence on base classes, e.g., \textit{Person/Horse/Sheep/Dog} in the $4^{th}$ column. 
This will result in mis-segmentation of those base classes into novel classes.
After the calibration by our cross-covariance Transformer, the novel classifier's overconfidence on base classes is clearly depressed.

\paragraph{Ablation Study of Our Transformer $\mathcal{T}$}
In this experiment, we compare our cross-covariance Transformer with other calibration variants that are linear layer based and self-attention based.
Specifically, linear layer based calibration replaces our cross-covariance Transformer with one fully-connected layer; while the self-attention based calibration uses [$y_{all}$, $y_{all}$, $y_{all}$] as inputs for self-attention.
The comparison is shown in Table~\ref{tab:query_con}. 
From the results we can see that linear layer based variants are not comparable to NSF, indicating the difficulty in calibrating the prediction with a simple fully-connected layer. 
Meanwhile, we can see the residual connection is helpful in prediction calibration.
Self attention also works well in this setup showing a Transformer architecture is beneficial. But it is much inferior to our proposed cross-covariance Transformer, showing the prediction and feature channel covariance is helpful in the prediction calibration in \textit{GFSS}.

\begin{figure}[ht]
    \centering
    \includegraphics[width=0.48\textwidth]{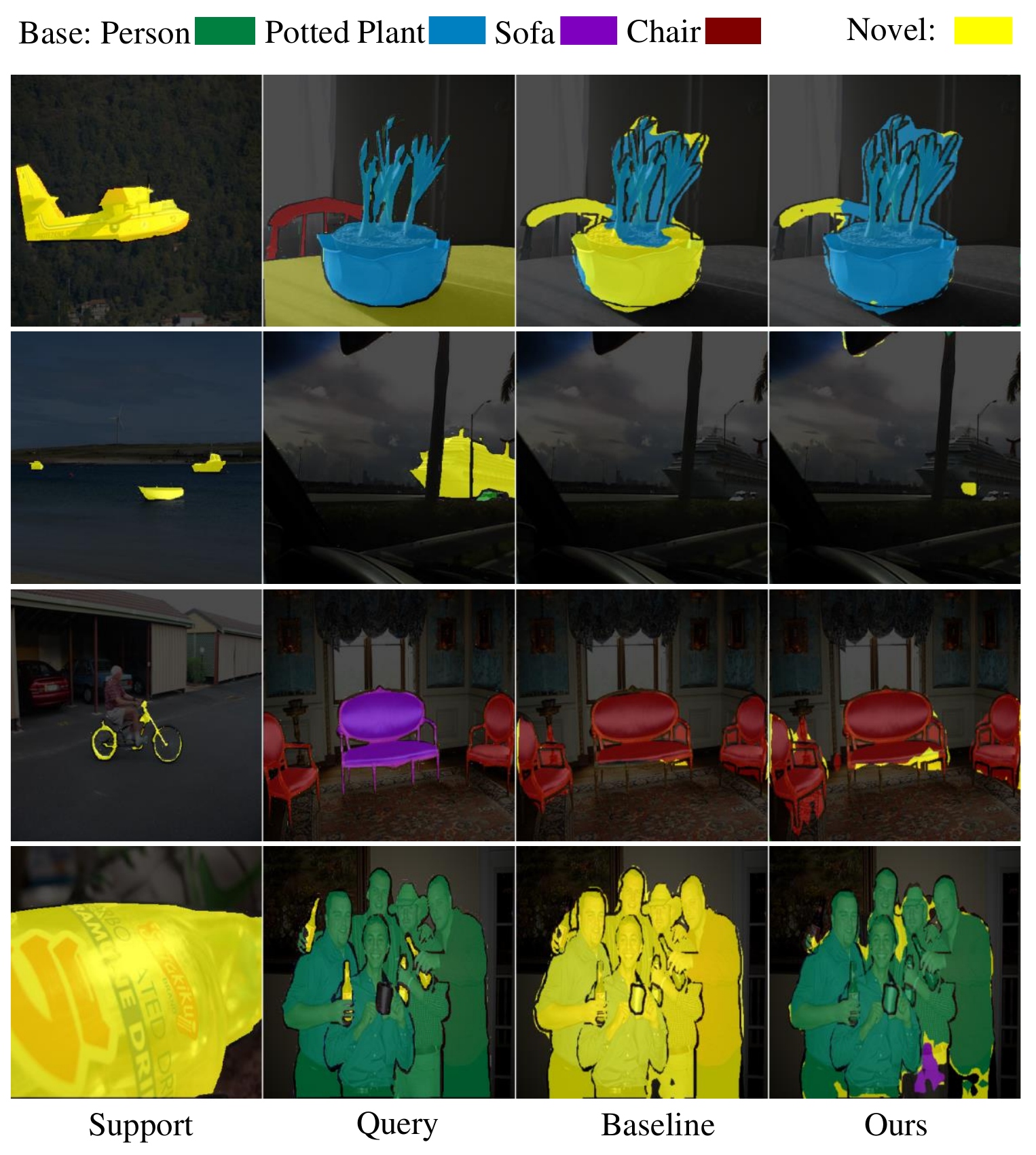}
    \caption{Failure cases. Dataset: PASCAL-$5^0$. Backbone: ResNet-$50$.}
    \label{fig:failure}
\end{figure}

\begin{table}[ht]
    \centering
    \caption{Ablation study of our proposed Transformer.  Dataset: PASCAL-5$^{0}$. Backbone: ResNet-50. $\dag$ means without residual connection.}
    \begin{tabular}{l|c|cccc}
    \hline
         \multirow{2}{*}{Calibration method} & \multirow{2}{*}{Shot} & \multicolumn{4}{c}{s-0} \\
         \cline{3-6}
          & & Base & Novel & mIoU & $H_{mean}$ \\
         \hline
          NSF & \multirow{5}{*}{1} & 29.38 & \textbf{51.86} & 35.00 & 37.51 \\
          Linear layer$^\dag$ & & 19.78 & 15.71 & 18.76 & 17.51 \\
          Linear layer & & 22.53 & 18.27 & 21.47 & 20.18 \\
          Self-attention & & \textbf{61.96} & 35.54 & 55.36 & 45.17 \\
          PCN (Ours) & & 59.43 & 47.98 & \textbf{56.57} & \textbf{53.09} \\
    \hline
    \end{tabular}
    \label{tab:query_con}
\end{table}

\paragraph{Comparisons among Varying-level Features Used for Cross-attention}
Table \ref{tab:feat_effect} shows the results using varying-level features for cross-attention.
Note that the high-level feature is produced by Pyramid Pooling Module (PPM) (bins [1,2,3,6]) from PSPNet \cite{zhao2017pyramid} and concatenation without combining low-level features.
The first observation is that in most cases low-level features work better than high-level features.
This is reasonable as low-level features contain more detailed information useful for segmentation.
Second, the combined features by both low-level and high-level features outperform all other variants in that it benefits from both detailed structure and abstract semantics.

\begin{table}[ht]
    \centering
    \caption{Comparisons among varying-level features used for cross-attention. Dataset: PASCAL-5$^{0}$. Backbone: ResNet-50.}
    \begin{adjustbox}{max width=0.48\textwidth}
    \begin{tabular}{l|c|ccccc}
    \hline
         \multirow{2}{*}{Feature} & \multirow{2}{*}{Shot} & \multicolumn{4}{c}{s-0} \\
         \cline{3-6}
          & & Base & Novel & mIoU & $H_{mean}$ \\
         \hline
         $Layer_2$ (Low-level) & \multirow{5}{*}{1} & 60.80 & 40.29 & 55.68 & 48.47 \\
         $Layer_3$ (Low-level) & & 60.94 & 37.42 & 55.06 & 46.37 \\
         $Layer_4$ (Low-level) & & 57.28 & 46.46 & 54.57 & 51.30 \\
         High-level & & \textbf{61.51} & 39.57 & 56.03 & 48.16 \\
         $Layer_4$ + High-level (Ours) & & 59.43 & \textbf{47.98} & \textbf{56.57} & \textbf{53.09} \\
    \hline
    \end{tabular}
    \end{adjustbox}
    \label{tab:feat_effect}
\end{table}

\paragraph{Comparisons of Computational Time}
Table \ref{tab:compute} shows the comparisons on training time and memory cost.
It seems the proposed method has comparable training time and memory cost which are merely inferior to the best competitors, but a much better $H_{mIoU}$ performance with 9.17\% gain.
Note that after training, there are no significant differences of the inference time of all compared methods.

\begin{table}[ht]
    \centering
    \caption{Comparing computational time and memory cost. Dataset: PASCAL-$5^0$. Backbone: ResNet-$50$. Batch size: $8$. $\dag$: exclude the base classifier pre-training time. The best results are in \textbf{bold} while the second best is \underline{underlined}. Results were obtained on one single NVIDIA GeForce RTX $2080$ Ti GPU.}
    \label{tab:compute}
    \begin{tabular}{l|c|ccc}
    \hline
         \multirow{2}{*}{Method} & \multirow{2}{*}{Shot} & \multicolumn{3}{c}{s-0} \\
         \cline{3-5}
         & & Training time$^{\dag}$ $\downarrow$ & Memory cost $\downarrow$ & $H_{mIoU} \uparrow$  \\
         \hline
         DENet & \multirow{4}{*}{1} & 31.4h & \textbf{4.2G} & 39.15 \\
         PFENet & & 70.5h & 8.6G & 21.26 \\
         ACASTLE & & \textbf{4.5h} & 5.7G & \underline{43.92} \\
         PCN (Ours) & & \underline{6.8h} & \underline{4.7G} & \textbf{53.09}  \\
    \hline
    \end{tabular}
\end{table}

\paragraph{Failure Cases}
Figure \ref{fig:failure} shows some failure cases. This can help better understand the limitation of the proposed method and offer insights on future works. Specifically, we observed that failure mainly occurs in three scenarios. First, when only part of the object is shown in the figure, e.g., $Chair$ in the $1^{st}$ row, the proposed \textit{PCN} tends to mis-segment that part to a class with a similar appearance. Second, when huge intra-class variances exist between support and query sample as shown in $2^{nd}$ and $4^{th}$ rows, the target objects are often missed. This is because the large intra-class variance makes the model hard to recognize them as the same class. Finally, the ambiguous annotations cause wrong segmentations. For example, the $Sofa$ in $3^{rd}$ row is visually more similar to the $Chair$ objects so that it is reasonable for \textit{PCN} to segment it to $Chair$ class. For the first two factors, we can solve them by modeling the relationship between the part and the global, and closing the representations within the same class. In contrast, the ambiguous annotations are needed to rectify.

\section{Conclusion}
We have proposed a novel  Prediction Calibration Networks (\textit{PCN}) for generalized few-shot semantic segmentation. 
It consists of two key modules: Normalized Score Fusion (NSF) and Transformer based prediction calibration.
NSF alone serves as a strong baseline that can mitigate the prediction bias towards base classes, but is prone to novel class bias. 
A new feature-score cross-covariance based Transformer is proposed to further calibrate prediction score and episodically learn to avoid any bias.
Extensive experiments on two standard benchmarks show our proposed framework achieves the state-of-the-art performance.
We also conducted comprehensive further analysis for better understanding how each component contributes to the final performance.



\bibliographystyle{IEEEtran}
\bibliography{ref.bib}


 

\begin{IEEEbiography}[{\includegraphics[width=1in,height=1.25in,clip,keepaspectratio]{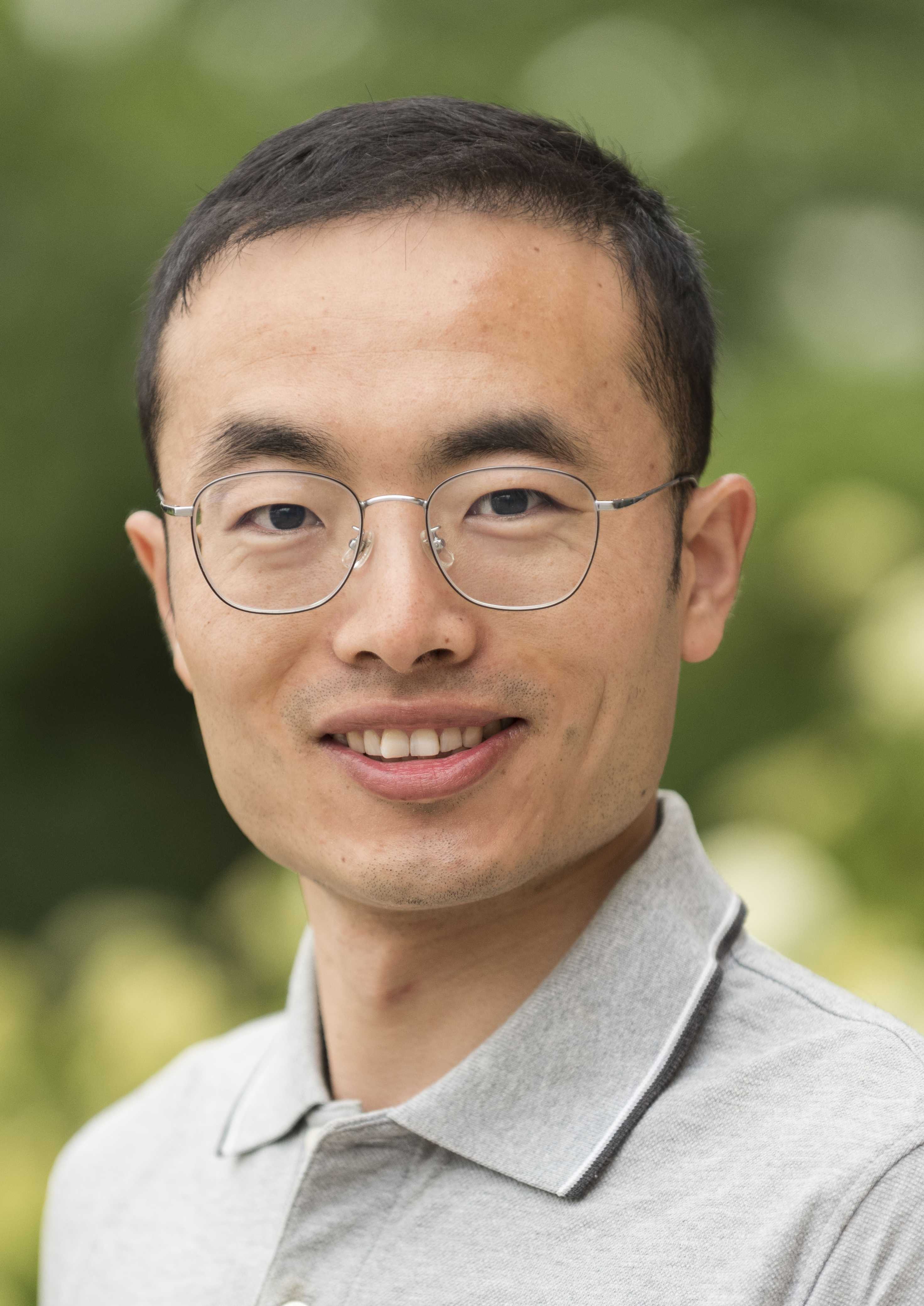}}]{Zhihe Lu} is now a Research Fellow at National University of Singapore. He received the Ph.D. degree at Centre for Vision, Speech and Signal Processing (CVSSP), University of Surrey in 2022, and the master degree at Chinese Academy of Sciences, Institute of Automation (CASIA) in 2019. His research interest centers around deep learning with limited annotated data. Recently, he focuses on two applications including domain adaptation and few-shot learning.
\end{IEEEbiography}

\begin{IEEEbiography}[{\includegraphics[width=1in,height=1.25in,clip,keepaspectratio]{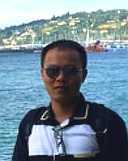}}]{Sen He}
received the Ph.D. degree at University of Exeter. He is now a Research Scientist at Meta AI, London. Previously, he was a Postdoctoral researcher at University of Surrey. He is broadly interested in the field of Computer Vision and Deep Learning. His current research focuses on Generative Models.
\end{IEEEbiography}

\begin{IEEEbiography}[{\includegraphics[width=1in,height=1.25in,clip,keepaspectratio]{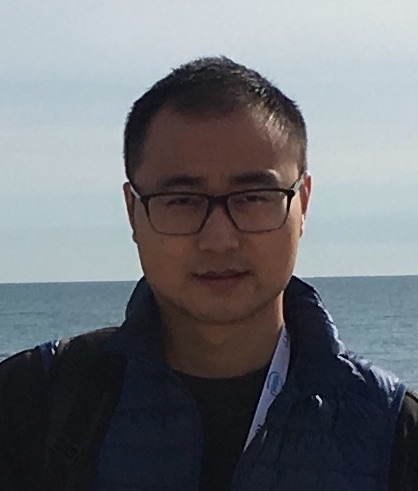}}]{Da Li}
is a Senior Research Scientist within the Machine Learning and Data Intelligence group in Samsung AI Centre Cambridge and a Visiting Researcher of the Machine Intelligence Research group at the University of Edinburgh. His research interests span transfer learning, meta-learning and semi-supervised learning. He serves as regular reviewers for flagship venues, e.g. CVPR, ICML, NeurIPS and journals, e.g. TPAMI, JMLR, ML, and has served as a Senior Program Committee (area chair) of AAAI 2022.
\end{IEEEbiography}

\begin{IEEEbiography}[{\includegraphics[width=1in,height=1.25in,clip,keepaspectratio]{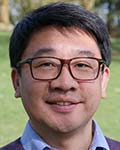}}]{Yi-Zhe Song}
(Senior Member, IEEE) received the bachelor’s degree (Hons.) from the University of Bath in 2003, the M.Sc. degree from the University of Cambridge in 2004, and the Ph.D. degree in computer vision and machine learning from the University of Bath in 2008. He is currently a Professor of computer vision and machine learning and the Director of the SketchX Laboratory, Centre for Vision Speech and Signal Processing (CVSSP), University of Surrey. Previously, he was a Senior Lecturer at the Queen Mary University of London and a Research and Teaching Fellow at the University of Bath. He is a fellow of the Higher Education Academy, as well as a full member of the EPSRC Review College, the UK’s main agency for funding research in engineering and the physical sciences. He received the Best Dissertation Award for his M.Sc. degree. He is the Program Chair of the British Machine Vision Conference (BMVC) in 2021 and regularly serves as the Area Chair (AC) for flagship computer vision and machine learning conferences, most recently at ICCV’21 and BMVC’20.
\end{IEEEbiography}

\begin{IEEEbiography}[{\includegraphics[width=1in,height=1.25in,clip,keepaspectratio]{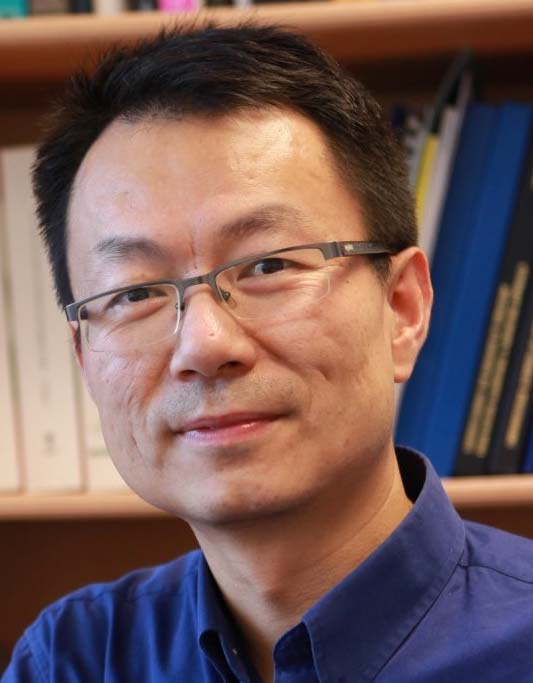}}]{Tao Xiang}
received the Ph.D. degree in electrical and computer engineering from the National University of Singapore in 2001. He is currently a Professor with the Department of Electrical and Electronic Engineering, University of Surrey, and a Research Scientist at Facebook AI. He has published over 200 papers in international journals and conferences. His research interests include computer vision, machine learning, and data mining.
\end{IEEEbiography}

\vfill

\end{document}